\documentclass[runningheads]{llncs}
\usepackage{times}
\usepackage{graphicx} % more modern
\usepackage[numbers]{natbib}
\usepackage{algorithm}
\usepackage{algorithmic}

\usepackage{multirow}

\newcommand{\us}{\ensuremath{u^\ast}}
\newcommand{\vs}{\ensuremath{v^\ast}}

\usepackage{amssymb}%,amsthm}%,mathtools}

\begin{document}
\title{
Machine vs Machine: Minimax-Optimal\\
Defense Against Adversarial Examples
}
\author{Jihun Hamm%\orcidID{0000-1111-2222-3333}
 \and
Akshay Mehra%\orcidID{1111-2222-3333-4444}
}
\authorrunning{J. Hamm and A. Mehra}
\institute{Dept. of Computer Science and Engineering,
The Ohio-State University, USA\\
\email{\{hamm.95,mehra.42\}@osu.edu}
%\url{http://wweb}
 }
\maketitle              % typeset the header of the contribution
\begin{abstract}
Recently, researchers have discovered that the state-of-the-art object classifiers
can be fooled easily by small perturbations in the input unnoticeable to human eyes. 
It is also known that an attacker can generate strong adversarial examples if she knows
the classifier parameters. Conversely, a defender can robustify the classifier
by retraining if she has access to the adversarial examples. 
We explain and formulate this adversarial example problem as a two-player
continuous zero-sum game, and demonstrate the fallacy of evaluating a defense or an attack 
as a static problem.
To find the best worst-case defense against whitebox attacks, we propose a 
continuous minimax optimization algorithm.
We demonstrate the minimax defense with two types of attack classes --gradient-based
and neural network-based attacks. 
Experiments with the MNIST and the CIFAR-10 datasets demonstrate that the defense found 
by numerical minimax optimization is indeed more robust than non-minimax defenses.
We discuss directions for improving the result toward achieving robustness
against multiple types of attack classes.
\end{abstract}

%%%%%%%%%%%%%%%%%%%%%%%%%%%%%%%%%%%%%%%%%%%%%%%%%%%%%%%%%%%%%%%%%%%%%%%%%%%%%%%%
\section{Introduction}
%%%%%%%%%%%%%%%%%%%%%%%%%%%%%%%%%%%%%%%%%%%%%%%%%%%%%%%%%%%%%%%%%%%%%%%%%%%%%%%%

Recently, researchers have made an unexpected discovery that the state-of-the-art
object classifiers can be fooled easily by small perturbations in the input unnoticeable 
to human eyes \citep{szegedy2013intriguing,goodfellow2014explaining}.
Following studies tried to explain the cause of the seeming failure of deep learning
toward such adversarial examples. The vulnerability was ascribed to linearity \citep{goodfellow2014explaining},
low flexibility \citep{fawzi2015analysis}, or the flatness/curvedness of decision boundaries
\citep{moosavi2017analysis}, but a more complete picture is still under research. 
This is troublesome since such a vulnerability can be exploited in critical situations
such as an autonomous car misreading traffic signs or a facial recognition system 
granting access to an impersonator without being noticed. 
% Attack & defense
Several methods of generating adversarial examples were proposed \citep{goodfellow2014explaining,moosavi2016universal,carlini2017towards},
most of which use the knowledge of the classifier to craft examples. 
In response, a few defense methods were proposed: 
retraining target classifiers with adversarial examples called adversarial training
\citep{szegedy2013intriguing,goodfellow2014explaining}; suppressing gradient by retraining
with soft labels called defensive distillation \citep{papernot2016distillation};
hardening target classifiers by training with an ensemble of adversarial examples \citep{tramer2017ensemble}. (See Related work for descriptions of more methods.)

In this paper we focus on {whitebox} attacks where
the model and the parameters of the classifier are known to the attacker.
This requires a genuinely robust classifier or defense method since the defender
cannot rely on the secrecy of the parameters as defense. 
To emphasize the dynamic nature of attack-defense, we start with the following simple 
experiments (see Sec~\ref{sec:motivating} for a full description).
Suppose first a classifier is trained on a original non-adversarial dataset. 
Using the trained classifier parameters, an attacker can then generate 
adversarial examples, e.g., using the fast gradient sign method (FGSM) \citep{goodfellow2014explaining} which is known to be simple and effective.
However, if the defender/classifier\footnote{The defender and the classifier
 are treated synonymously in this paper.} has access to those {adversarial examples}, 
the defender can significantly weaken the attack by retraining the classifier with 
adversarial examples, called {adversarial training}. 
We can repeat the two steps -- adversarial sample generation and adversarial training --
many times, and what is observed in the process (Sec~\ref{sec:motivating}) is that 
an attack/defense can be very effective against immediately-preceding defense/attack,
but not necessarily against non-immediately preceding defenses/attacks. 
This is one of many examples that show that the effectiveness of an attack/defense method
depends critically on the defense/attack it is against, 
from which we conclude that the performance of an attack/defense method have to be evaluated
and reported in the attack-defense pair and not in isolation.

% game formulation
To better understand the interaction of attack and defense in the adversarial example problem,
we formulate adversarial attacks/defense on machine learning classifiers 
as a two-player continuous pure-strategy zero-sum game. 
The game is played by an attacker and a defender where the attacker tries to maximize
the risk of the classification task by perturbing input samples under certain constraints,
and the defender tries to adjust the classifier parameters to minimize the same risk function
given the perturbed inputs. 
% Two classes
The ideal adversarial examples are the global maximizers of the risk without
constraints (except for certain bounds such as the $l_p$-norm.)
However, such a space of unconstrained adversarial samples is very large -- any real vector
of the given input size is potentially an adversarial sample regardless of whether it is
a sample from the input data distribution. 
The vastness of the space of adversarial examples is a hindrance to the study
of the problem, since it is difficult for the defender to model and learn the attack class
from a finite number of adversarial examples and generalize to future attacks. 
To study the problem more concretely, we use two representative classes of attacks.
The first type is of gradient-type -- the attacker uses mainly the gradient of the classifier
output with respect to the input to generate adversarial examples. 
This includes the fast gradient sign method (FGSM) \citep{goodfellow2014explaining} 
and the iterative version (IFGSM) \citep{kurakin2016adversarial} of the FGSM. 
Attacks of this type can be considered an approximation of the full maximization of the risk
by one or a few steps of gradient-based maximization. 
The second type is a neural-network based attack which is capable of learning.
This attack network is trained from data so that it take a (clean) input and generate 
a perturbed output to maximally fool the classifier in consideration. 
The `size' of the attack class is directly related to the parameter space of 
the neural network architecture,
e.g., all perturbations that can be generated by fully-connected 3-layer ReLU networks
that we use in the paper. 
Similar to what we propose, others have recently considered training neural networks  
to generate adversarial examples \citep{nguyen2017learning,baluja2017adversarial}. 
While the network-based attack is a subset of the space of unconstrained attacks,
it can generate adversarial examples with only a single feedforward pass through the neural
network during test time,
making it suitable for real-time attacks unlike other more time-consuming attacks. 
We later show that this class of neural-network based attacks is quite different from the
the class of gradient-based attacks empirically.

% Equilibrium and minimax, and minimax algorithm
As a two-player game, there may not be a dominant defense that is robust against all types of
attacks. 
However, there is a natural notion of the best defense or attack in the worst case. 
Suppose one player moves first by choosing her parameters and the other player responds
with the knowledge of the first player's move. This is an example of 
a leader-follower game \citep{bruckner2011stackelberg} for which there are two
known equilibria -- the minimax and the maximin points if it is a constant-sum game. 
Such a defense/attack is theoretically an ideal pure strategy in the leader-follower setting,
but one has to actually find it for the given class of defense/attack and 
the dataset to deploy the defense/attack.
To find minimax solutions numerically, we propose continuous optimization algorithms
for the gradient-based attacks (Sec.~\ref{sec:minimax-grad}) and the network-based attacks
(Sec.~\ref{sec:minimax-attnet}), based on alternating minimization with gradient-norm penalization. 
Experiments with the MNIST and the CIFAR-10 datasets 
show that the minimax defense found from the algorithm is indeed more robust than 
non-minimax defenses overall including adversarially-trained classifiers against
specific attacks. 
However, the results also show that the minimax defense is still vulnerable to some degrees
to attacks from out-of-class attacks, e.g., the gradient-based minimax defense is
not equally robust against network-based attacks. 
This exemplifies the difficulty of achieving the minimax defense against all 
possible attack types in reality.
Our paper is a first step towards this goal, and future works are discussed in Sec.~\ref{sec:discussion}.

The contributions of this paper can be summarized as follows.
\begin{itemize}
\item We explain and formulate the adversarial example problem as a two-player continuous game,
and demonstrate the fallacy of evaluating a defense or an attack as a static problem. 
\item We show the difficulty of achieving robustness against all types of attacks,
and present the minimax defense as the best worst-case defense.
\item We present two types of attack classes -- gradient-based and network-based attacks.
The former class represents the majority of known attacks in the literature,
and the latter class represents new attacks capable of generating adversarial examples 
possessing very different properties from the former. 
\item We provide continuous minimax optimization methods to find the minimax point for the two
classes, and contrast it with non-minimax approaches.
\item We demonstrate our game formulation using two popular machine learning benchmark
datasets, and provide empirical evidences to our claims.
\end{itemize}

For readability, details about experimental settings and the results with the
CIFAR-10 dataset are presented in the appendix. 
%To facilitate the replication of the paper, the codes used to generate the results in this paper will appear in a public code repository.%\footnote{\url{github.com/...}}.

%%%%%%%%%%%%%%%%%%%%%%%%%%%%%%%%%%%%%%%%%%%%%%%%%%%%%%%%%%%%%%%%%%%%%%%%%%%%%%%%
\section{Related work}
%%%%%%%%%%%%%%%%%%%%%%%%%%%%%%%%%%%%%%%%%%%%%%%%%%%%%%%%%%%%%%%%%%%%%%%%%%%%%%%%

Making a classifier robust to test-time adversarial attacks
has been studied for linear (kernel) hyperplanes \citep{lanckriet2002robust}, 
naive Bayes \citep{dalvi2004adversarial} and SVM \citep{globerson2006nightmare},
which also showed the game-theoretic nature of the robust classification problems.
Since the recent discovery of adversarial examples for deep neural networks, 
several methods of generating adversarial samples were proposed \citep{szegedy2013intriguing,goodfellow2014explaining,huang2015learning,moosavi2016universal,
carlini2017towards}
as well as several methods of defense \citep{szegedy2013intriguing,goodfellow2014explaining,
papernot2016distillation,tramer2017ensemble}. 
These papers considered static scenarios, where the attack/defense
is constructed against a fixed opponent. 
A few researchers have also proposed using a detector to detect and reject
adversarial examples \citep{meng2017magnet,lu2017safetynet,metzen2017detecting}.
While we do not use detectors in this work, the minimax approach we proposed in the paper
can be applied to train the detectors. 

The idea of using neural networks to generate adversarial samples has appeared concurrently \citep{baluja2017adversarial,nguyen2017learning}. Similar to our paper,
the two papers demonstrates that it is possible to generate strong adversarial samples by 
a learning approach. The former \citep{baluja2017adversarial} explored different architectures for
the ``adversarial transformation networks'' against several different classifiers.
The latter \citep{nguyen2017learning} proposed ``attack learning neural networks'' to map 
clean samples to a region in the feature space where misclassification occurs and
``defense learning neural networks'' to map them back to the safe region. 
Instead of prepending the defense layers before the fixed classifier \citep{nguyen2017learning}, we retrain 
the whole classifier as a defense method. 
However, the key difference of our work to the two papers is that we consider
the dynamics of a learning-based defense stacked with a learning-based attack,
and the numerical computation of the optimal defense/attack by continuous optimization.

Extending the model of \cite{huang2015learning}, an alternating optimization algorithm for 
finding saddle-points of the iterative FGSM-type attack and the adversarial training defense
was proposed recently in \citep{madry2017towards}. 
The algorithm is similar to what we propose in Sec.~\ref{sec:minimax-grad}, but the  
difference is that we seek to find minimax points instead of saddle points, as well as 
we consider both the gradient-based and the network-based attacks.
The importance of distinguishing minimax and saddle-point solutions for machine learning
problems was explained in \cite{hamm2018k-beam} along with new algorithms for handing multiple 
local optima. 
The alternating optimization method for finding an equilibrium of a game has 
gained renewed interest since the introduction of Generative Adversarial Networks (GAN)  \citep{goodfellow2014generative}. However, the instability of the alternating gradient-descent 
method has been known, and the ``unrolling'' method \citep{metz2016unrolled} was proposed
to speed up the GAN training.
The optimization algorithm proposed in the paper has a similarity with the unrolling method,
but it is simpler and involves a gradient-norm regularization 
which can be interpreted intuitively as sensitivity penalization \citep{gu2014towards,lyu2015unified,mescheder2017numerics,nagarajan2017gradient,roth2017stabilizing}.

Lastly, a related framework of finding minimax risk was also studied in \cite{hamm2017minimax}
for the purpose of preventing attacks on privacy. We discuss how the attack on classification 
in this paper and the attack on privacy are the two sides of the same optimization problem with 
the opposite goals.

%%%%%%%%%%%%%%%%%%%%%%%%%%%%%%%%%%%%%%%%%%%%%%%%%%%%%%%%%%%%%%%%%%%%%%%%%%%%%%%%
\section{Minimax defense against gradient-based attacks}\label{sec:minimax-grad}
%%%%%%%%%%%%%%%%%%%%%%%%%%%%%%%%%%%%%%%%%%%%%%%%%%%%%%%%%%%%%%%%%%%%%%%%%%%%%%%%

A classifier whose parameters are known to an attacker is easy to attack. 
Conversely, an attacker whose adversarial samples are known to a classifier 
is easy to defend against. 
In this section, we first demonstrate the cat-and-mouse nature of the interaction,
using adversarial training as defense
and the fast gradient sign method (FGSM) \citep{goodfellow2014explaining} attack.
We then describe more general form of the game, and algorithms for finding 
minimax solutions using sensitivity-penalized optimization. 

\subsection{A motivating observation}\label{sec:motivating}

Suppose $g$ is a classifier $g:\mathcal{X} \to \mathcal{Y}$
and $l(g(x),y)$ is a loss function. 
The untargeted FGSM attack generates a perturbed example $z(x)$ given the clean sample $x$
as follows:
\begin{equation}\label{eq:fgsm}
z(x) = x + \eta\; \mathrm{sign}\;(\nabla_x l(g(x),y)).
\end{equation}
Untargeted means that the goal of the attacker is to induce misclassification regardless
of which classes the samples are misclassified into, as long as they are different from 
the original classes. We will not discuss defense against targeted attacks in the paper
as they are analogous to untargeted counterparts in the paper. 
%We use the $l_\infty$ norm in this paper, in which case
%$z(x) = x + \eta\; \mathrm{sign}(\nabla_x l(g(x),y))$. 
The clean input images we use here are $l_\infty$-normalized, that is, 
all pixel values are in the range $[-1,1]$. 
\if0
It was argued that the use of true label $y$ results in ``label leaking'' 
\citep{kurakin2016adversarial2}, but we use will true labels in the paper for simplicity.
For another attack example, the IFGSM attack iteratively refines an adversarial example
by the following update
\begin{equation}
z_{i+1} = \mathrm{clip}_{x,\eta}(z_i + \eta\; \mathrm{sign}(\nabla_z l(g(z_i),y))),
\end{equation}
where the clipping used in this paper is 
$\mathrm{clip}_{x,\eta}(x') \triangleq \min\{1,\;x+\eta,\;\max\{-1,\;x-\eta,\;x'\}\}.$
\fi
Although simple, FGSM is very effective at fooling the classifier.
Table~\ref{tbl:known_defense_mnist} demonstrates this against 
a convolutional neural network trained with clean images from MNIST.
(Details of the classifier architecture and the settings are in the appendix.)

\begin{table}[htb]
\small
%\begin{center}
\renewcommand{\arraystretch}{0.9}
\centering
\begin{tabular}{|c||c|cccc|cccc|}
\hline
\multirow{2}{*}{Defense\textbackslash Attack} & \multirow{2}{*}{No attack} & 
\multicolumn{4}{c|}{FGSM}\\
\cline{3-6}
% & & $\eta$=0.2 & $\eta$=0.3 & $\eta$=0.4 & $\eta$=0.5 \\
%\hline
%No defense & 0.026 & 0.933 & 0.983 & 0.985 & 0.987 \\
 & & $\eta$=0.1 & $\eta$=0.2 & $\eta$=0.3 & $\eta$=0.4\\
\hline
No defense & 0.026 & 0.446 & 0.933 & 0.983 & 0.985 \\
\hline
\end{tabular} 
\caption{Test error rates of the FGSM attack on an undefended convolutional neural network for MNIST. Higher error means a more successful attack.}
\label{tbl:known_defense_mnist}
%\end{center}
\end{table}

\vspace{-0.3in}
On the other hand, these attacks, if known to the classifier,
can be weakened by retraining the classifier with the original dataset
augmented by adversarial examples with ground-truth labels, 
known as adversarial training. 
In this paper we use the $1:1$ mixture of the clean and the adversarial samples 
for adversarial training.
Table~\ref{tbl:known_attack_mnist} shows the result of adversarial training
for different values of $\eta$. 
After adversarial training, the test error rates for adversarial test examples are reduced
back to the level (1-2\%) before attack. This is in stark contrast with the high
 misclassification of the undefended classifier in Table~\ref{tbl:known_defense_mnist}.

\begin{table}[htb]
\small
%\begin{center}
\renewcommand{\arraystretch}{0.9}
\centering
\begin{tabular}{|c||c|cccc|}
\hline
\multirow{2}{*}{Defense\textbackslash Attack} & \multirow{2}{*}{No attack} & 
\multicolumn{4}{c|}{FGSM}\\
\cline{3-6}
% & & $\eta$=0.2 & $\eta$=0.3 & $\eta$=0.4 & $\eta$=0.5 \\
%\hline 
%Adv train & n/a & 0.011 & 0.015 & 0.017 & 0.021\\
 & & $\eta$=0.1 & $\eta$=0.2 & $\eta$=0.3 & $\eta$=0.4 \\
\hline 
Adv train & n/a & 0.010 & 0.011 & 0.015 & 0.017\\
\hline
\end{tabular} 
\caption{Test error rates of the FGSM attacks on adversarially-trained classifiers for MNIST. 
This defense can avert the attacks and achieve low error rates.}
\label{tbl:known_attack_mnist}
%\end{center}
\end{table}

\vspace{-0.3in}
This procedure of 1) adversarial sample generation using the current classifier,
and 2) retraining classifier using the current adversarial examples, can be repeated 
for many rounds. 
%The answer to this cat-and-mouse game is easy to experiment although time-consuming. 
Let's denote the attack on the original classifier as FGSM1, and the corresponding
retrained classifier as Adv FGSM1. 
Repeating the procedure above generates the sequence of attacks and defenses:
 FGSM1 $\to$ Adv FGSM1 $\to$ FGSM2 $\to$ Adv FGSM2 $\to$ FGSM3 $\to$ Adv FGSM3, etc.
The odd terms are attacks and the even terms are defenses (i.e., classifiers.) 
\if0
Fig.~\ref{fig:cat-and-mouse_mnist} shows one such trial with 80 + 80 rounds of the procedure.
Initially, the attacker achieves near-perfect attacks (i.e., error rate $\simeq$ 1), 
and the defender achieves near-perfect defense (i.e., error rate $\simeq$ 0).
As the iteration increases, the attacker becomes weaker with error rate$\simeq$ 0.5,
but the defense is still very successful, and the rate seems to oscillate persistently.
% Why the convergence to small error is faster with larger $\eta$'s? Shouldn't it be the opposite?
While we can run more iterations to see if it converges, 
this is not a very principled nor efficient approach to find an equilibrium,
if it exists. 
\fi

We repeat this two step for 80 rounds. 
As a preview, Table~\ref{tbl:normtrain_mnist} shows the test errors of the defense-attacks pairs
where the defense is one of the \{No defense, Adv FGSM1, Adv FGSM2, ...\} and the attacker
is one of the \{No attack, FGSM1, FGSM2, ...\}. 
Throughout the paper we use the following conventions for tables: 
the rows correspond to defense methods and the columns correspond to attack methods.
All numbers are test errors.
It is observed that the defense is effective against the immediately-preceding defense
(e.g., Adv FGSM1 defense has a low error against FGSM1 attack), and 
similarly the attack is effective against the immediately-preceding defense 
(e.g., FGSM2 attack has a high error against Adv FGSM1).
However, a defense/attack is not necessarily robust against other non-immediately preceding 
attack/defense. 
From this we make two observations. 
%This observation sends us a warning as to how we evaluate the performance of an attack method  %or a defense. 
First, the effectiveness of an attack/defense method depends critically on the defense/attack
it is against, and therefore the performance of an attack/defense should be evaluated
as the attack-defense pair and not in isolation. 
Second, it is not enough for a defender to choose the classifier parameters in response
to a specific attack., i.e., adversarial training, but it should use a more principled method
of selecting robust parameters. We address these below.

\if0
\begin{figure*}[thb]
\centering
%This is a placeholder.
\includegraphics[width=1\linewidth]{figures/robustnet_catmouse_eta030_mnist.pdf}
\caption{A cat-and-mouse game of FGSM attacks and adversarial training for MNIST.
The lower green points are the error rates after adversarial training, 
and the upper red points are the error rates after FGSM attack ($\eta=0.3$). 
After 160 iterations, the error rate is still oscillating between 0 and 0.5.
}
\label{fig:cat-and-mouse_mnist}
\end{figure*}
\fi

\subsection{Gradient-based attacks and generalization}

We first consider the interaction of the classifier and the gradient-based attack
as a continuous two-player pure-strategy zero-sum game.
To emphasize the parameters $u$ of the classifier/defender $g(x;u)$, let's write the
empirical risk of classifying the perturbed data as 
\begin{equation}\label{eq:f}
f(u,Z) \triangleq \frac{1}{N}\sum_{i=1}^N l(g({z}(x_i);u),y_i),
\end{equation}
where ${z}(x)$ denote a gradient-based attack based on the loss gradient
\begin{equation}\label{eq:gradient-based}
{z}(x) \leftarrow x + \eta\;\nabla_x l(g(z;u),y),
\end{equation}
and $Z=(z_1,...,z_N)\triangleq(z(x_1),...,z(x_N))$ is the sequence of perturbed examples. 

%\subsection{Generalization}
Given the classifier parameter $u$, the gradient-based attack (Eq.~\ref{eq:gradient-based})
can be considered as a single-step approximation to the general attack 
\begin{equation}
\max_Z f(u,Z)
\end{equation}
where $Z=(z_1,...,z_N)$ can be any adversarial pattern subject to bounds such as
 $\|z_i\|_\infty\leq 1$ and $\|z_i - x_i\|_p \leq \eta$.
Consequently, the goal of the defender is to choose the classifier parameter $u$ to 
minimize the maximum risk from such attacks \cite{huang2015learning,madry2017towards}
\begin{equation}\label{eq:minimax}
\min_u \max_Z f(u,Z),
\end{equation}
with the same bound constraints. 
This general minimax optimization is difficult to solve directly
due to the large search space $Z$ of the inner maximization. 
In this respect, existing attack methods such as FGSM, IFGSM, or 
Carlini-Wagner \citep{carlini2017towards}
can be consider as heuristics or approximations of the true maximization. 

\subsection{Minimax solutions}

%Let us write the expected or empirical loss $L(\hat{z},y;u)$ of the classifier as 
%a function $f(u,z) = L(\hat{z},y;u)$ to emphasize the two main variables: 
%the defense $u$ and the perturbed input $\hat{z}$.
%With the abuse of the notation, here $z$ is also the perturbed input to the classifier/defender. 
We describe an algorithm to find the solutions of Eq.~\ref{eq:minimax} for
gradient-based attacks (Eq.~\ref{eq:gradient-based}). 
In expectation of the attack, the defender should choose $u$ to minimize
$f(u,{Z}(u))$ where the dependence of the attack on the classifier $u$ is
expressed explicitly. 
%If we can minimize $f(u,{z}(u))$ directly, we can find a stable defense/classifier
%without performing the repeated procedure of FGSM followed by adversarial training.
If we minimize $f$ using gradient descent 
\begin{equation}\label{eq:gd1}
u \leftarrow u - \lambda \frac{df(u,Z)}{du},
\end{equation}
then from the chain rule, the total derivative $\frac{df}{du}$ is
\begin{equation}\label{eq:grad1}
\frac{df}{du} 
= \frac{\partial f}{\partial u} + \frac{\partial {Z}}{\partial u} \frac{\partial f}{\partial Z}
= \frac{\partial f}{\partial u} + \sum_i \frac{\partial {z_i}}{\partial u} \frac{\partial f}{\partial z_i}
=\frac{\partial f}{\partial u} + \frac{\eta}{N}\sum_i \frac{\partial^2 l}{\partial x_i \partial u} \frac{\partial l}{\partial x_i}
\end{equation}
from Eqs.~\ref{eq:f} and \ref{eq:gradient-based}.

Interestingly, this total derivative (Eq.~\ref{eq:grad1}) at the current state
coincides with the gradient $\nabla_u$ of the following cost
\begin{equation}\label{eq:sensitivity}
\tilde{f}(u) \triangleq f(u,Z) + \frac{\gamma}{2} \left\|\frac{\partial f(u,Z)}{\partial Z}\right\|^2
=f(u,Z)+\frac{\eta}{2N} \sum_{i=1}^N \left\|\frac{\partial l(g(z_i;u),y_i)}{\partial z_i}\right\|^2
\end{equation}
where $\gamma=\eta N$.
There are two implications. Interpretation-wise, this cost function is the sum
of the original risk $f$ and the `sensitivity' term $\|\partial f/\partial Z\|^2$
which penalizes abrupt changes of the risk w.r.t. the input. 
Therefore, $u$ is chosen at each iteration to not only decrease the risk
but also to make the classifier insensitive to input perturbation so that the 
attacker cannot take advantage of large gradients.
The idea of minimizing the sensitivity to input is a familiar approach 
in robustifying classifiers \citep{gu2014towards,lyu2015unified}.
Secondly, the new formulation can be implemented easily. 
The gradient descent update using the 
seemingly complicated gradient (Eq.~\ref{eq:grad1}) can be replaced by the gradient descent
update of Eq.~\ref{eq:sensitivity}.
The capability of automatic differentiation \citep{rall1981automatic} in
modern machine learning libraries can be used to compute the gradient of Eq.~\ref{eq:sensitivity} efficiently. 

We find the solution to the minimax problem by iterating the two steps. 
In the max step, we generate the current adversarial pattern  
$z_i \leftarrow x_i + \eta \nabla_x l$, and in the min step, we update the
classifier parameters by $u \leftarrow u -\lambda \nabla_u \tilde{f}(u)$ using Eq.~\ref{eq:sensitivity}.
In practice, we require the adversarial patterns $z_i$'s have to be constrained by 
$\|z_i\|_\infty\leq 1$ and $\|z_i - x_i\|_p \leq \eta$, and therefore we can use the FGSM
method (Eq.~\ref{eq:fgsm}) to generate the patterns $z_i$'s in the max step.
The resultant classifier parameters $u$ after convergence will be referred to as minimax defense 
against gradient-based attacks ({\bf Minimax-Grad}). 
Note that this algorithm is similar but different from the algorithms of
\cite{huang2015learning,madry2017towards}. Firstly, we use only one gradient-descent step
to compute $Z$ although we can use multiple steps as in \cite{madry2017towards}. 
More importantly, the sensitivity penalty in Eq.~\ref{eq:sensitivity} plays an important
role for convergences in finding a minimax solution. 
In contrast, simply repeating $\min_u$ and $\max_Z$ without the penalty term does not guarantee
convergence to minimax points unless minimax points are also saddle points
(see \citep{hamm2018k-beam} for the description of the difference.) 
This subtle difference will be observed in the experiments. 

\subsection{Experiments}\label{sec:experiments1}

\begin{figure*}[thb]
\centering
%This is a placeholder.
\includegraphics[width=1\linewidth]{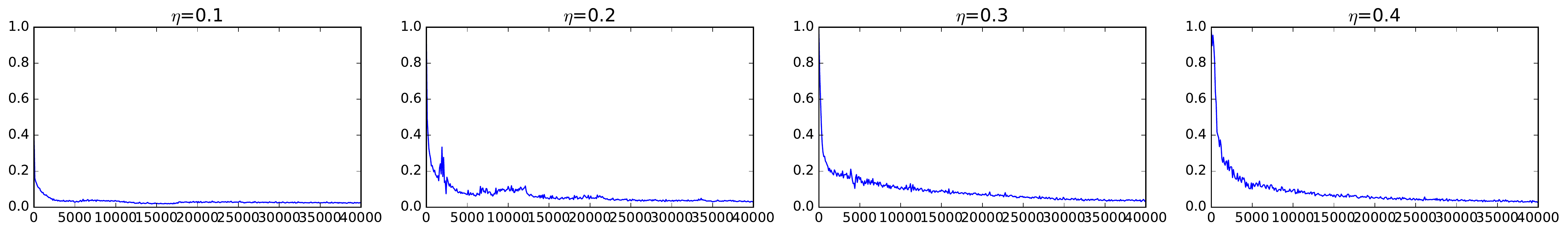}
\caption{Convergence of test error rates for Minimax-Grad with MNIST.
}
\label{fig:normtrain_mnist}
\end{figure*}

%in (\ref{eq:sensitivity}).
We find the defense parameters $u$ using the algorithm above, which will be robust to 
gradient-based attacks.
Fig.~\ref{fig:normtrain_mnist} shows the decrease of test error during training using 
this gradient descent approach for MNIST. 
%It only takes a very small fraction of time to reach the final states of
%the Fig.~\ref{fig:normtrain_mnist} compared to that of Fig.~\ref{fig:cat-and-mouse_mnist}. 

We reiterate the result of the cat-and-mouse game in Sec.~\ref{sec:motivating} and 
contrast it with the minimax solution (Minimax-Grad.)
Table~\ref{tbl:normtrain_mnist} shows that the adversarially trained classifier
(Adv FGSM1) is robust to both clean data and FGSM1 attack, 
but is susceptible to FGSM2 attack, showing that the defense is only effective against
immediately-preceding attacks. 
The same holds for Adv FGSM2, Adv FGSM3, etc. 
After 80 rounds of the cat-and-mouse procedure, the classifier Adv FGSM80
becomes robust to FGSM80 as well as moderately robust to other attacks including 
FGSM81 (=FGSM-curr). 
However, Minimax-Grad from the minimization of Eq.~\ref{eq:sensitivity}
is even more robust toward FGSM-curr than Adv FGSM80 and is overall the best. 
(See the last column -- ``worst'' result.)
To see the advantage of the sensitivity term in Eq.~\ref{eq:sensitivity},
we also performed the minimization of Eq.~\ref{eq:sensitivity} without the sensitivity term
under the same conditions as Minimax-Grad.  
This optimization method is similar to the method proposed in \cite{huang2015learning},
which we will refer to as LWA (Learning with Adversaries). 
Note that LWA is a saddle-point solution for gradient-based attacks since it 
solves $\min_u$ and $\max_Z$ symmetrically. 
In the table, one can see that Minimax-Grad is also better than LWA overall, 
although the difference is not large. 
%Note that Minimax-Grad is better than other adversarially-trained classifiers, but it too is
%still vulnerable to attacks such as FGSM80. 
To improve the minimax defense even further, we can choose a larger attack class
than single gradient-step attacks. Note that this will come at the cost of the
increased difficulty in minimax optimization. 
%This vulnerability raises the question if it is possible to make a classifier robust
%to any type of attacks, or more practically, robust to at least a large class of attacks.
%We discuss this issue in the next section. 

\begin{table}[htb]
\small
%\begin{center}
\renewcommand{\arraystretch}{0.9}
\centering
\begin{tabular}{|c|c||c|cccc|c|c|}
\cline{2-9}
\multicolumn{1}{c|}{}
&\multirow{2}{*}{Defense\textbackslash Attack} & \multirow{2}{*}{No attack} & 
\multicolumn{4}{|c|}{FGSM} & \multirow{2}{*}{FGSM-curr} & \multirow{2}{*}{worst} \\
\cline{4-7}
\multicolumn{1}{c|}{}& & &FGSM1 & FGSM2 & $\cdots$ & FGSM80 & & \\
\cline{2-9} 
\hline
\multirow{5}{*}{$\eta$=0.1}
& No defense   & 0.026 & 0.446 & 0.073 & $\cdots$ & 0.054 & 0.446 & 0.446 \\ 
&   Adv FGSM1  & 0.008 & 0.010 & 0.404 & $\cdots$ & 0.037 & 0.435 & 0.435 \\ 
&   Adv FGSM2  & 0.011 & 0.311 & 0.009 & $\cdots$ & 0.038 & 0.442 & 0.442 \\ 
&  Adv FGSM80  & 0.007 & 0.028 & 0.018 & $\cdots$ & 0.010 & 0.117 & 0.117 \\ 
&         LWA  & 0.009 & 0.044 & 0.030 & $\cdots$ & 0.022 & 0.019 & 0.044 \\
&Minimax-Grad  & 0.006 & 0.014 & 0.015 & $\cdots$ & 0.014 & 0.025 & 0.025 \\
\hline
\multirow{5}{*}{$\eta$=0.2}
&  No defense  & 0.026 & 0.933 & 0.215 & $\cdots$ & 0.089 & 0.933 & 0.933\\
&   Adv FGSM1  & 0.009 & 0.011 & 0.816 & $\cdots$ & 0.067 & 0.816 & 0.816 \\
&   Adv FGSM2  & 0.008 & 0.904 & 0.010 & $\cdots$ & 0.082 & 0.840 & 0.904 \\
&  Adv FGSM80  & 0.007 & 0.087 & 0.053 & $\cdots$ & 0.013 & 0.131 & 0.131 \\
&         LWA  & 0.007 & 0.157 & 0.034 & $\cdots$ & 0.036 & 0.026 & 0.157 \\
&Minimax-Grad  & 0.008 & 0.082 & 0.085 & $\cdots$ & 0.049 & 0.027 & 0.085 \\
\hline
\multirow{5}{*}{$\eta$=0.3}
&  No defense  & 0.026 & 0.983 & 0.566 & $\cdots$ & 0.087 & 0.983 & 0.983 \\
&   Adv FGSM1  & 0.010 & 0.015 & 0.892 & $\cdots$ & 0.080 & 0.892 & 0.892 \\
&   Adv FGSM2  & 0.010 & 0.841 & 0.017 & $\cdots$ & 0.058 & 0.764 & 0.841 \\
&  Adv FGSM80  & 0.007 & 0.352 & 0.117 & $\cdots$ & 0.021 & 0.043 & 0.352 \\
&         LWA  & 0.008 & 0.130 & 0.077 & $\cdots$ & 0.047 & 0.034 & 0.130 \\
&Minimax-Grad  & 0.008 & 0.062 & 0.144 & $\cdots$ & 0.045 & 0.036 & 0.144 \\
\hline
\multirow{5}{*}{$\eta$=0.4}
&  No defense  & 0.026 & 0.985 & 0.806 & $\cdots$ & 0.122 & 0.985 & 0.985 \\
&   Adv FGSM1  & 0.010 & 0.017 & 0.898 & $\cdots$ & 0.102 & 0.898 & 0.898 \\
&   Adv FGSM2  & 0.010 & 0.681 & 0.022 & $\cdots$ & 0.092 & 0.686 & 0.686 \\
&  Adv FGSM80  & 0.008 & 0.688 & 0.330 & $\cdots$ & 0.029 & 0.031 & 0.688 \\
&         LWA  & 0.009 & 0.355 & 0.171 & $\cdots$ & 0.086 & 0.042 & 0.355 \\
&Minimax-Grad  & 0.009 & 0.081 & 0.221 & $\cdots$ & 0.076 & 0.026 & 0.221 \\
\hline
%\multirow{5}{*}{$\eta$=0.5}
%&  No defense  & 0.026 & 0.987 & 0.856 & $\cdots$ & 0.470 & 0.987 & 0.987 \\
%&   Adv FGSM1  & 0.011 & 0.021 & 0.859 & $\cdots$ & 0.452 & 0.859 & 0.859 \\
%&   Adv FGSM2  & 0.009 & 0.595 & 0.032 & $\cdots$ & 0.411 & 0.495 & 0.595 \\
%&  Adv FGSM80  & 0.010 & 0.764 & 0.421 & $\cdots$ & 0.050 & 0.066 & 0.764 \\
%&         LWA  & 0.010 & 0.471 & 0.274 & $\cdots$ & 0.415 & 0.040 & 0.471 \\
%&Minimax-Grad  & 0.008 & 0.287 & 0.473 & $\cdots$ & 0.457 & 0.016 & 0.457 \\
%\hline
\end{tabular} 
\caption{Test error rates of different attacks on various adversarially-trained classifiers for MNIST. 
The rows correspond to defense methods and the columns correspond to attack methods.
FGSM-curr means the FGSM attack on the specific classifier on the left.
Worst means the largest error in each row. 
Adv FGSM is the classifier adversarially trained with FGSM attacks.
Minimax-Grad is the result of minimizing Eq.~\ref{eq:sensitivity} by gradient descent.
LWA is the result of minimizing Eq.~\ref{eq:sensitivity} without the gradient-norm term.
}
\label{tbl:normtrain_mnist}
%\end{center}
\end{table}

%\vspace{-0.1in}
%%%%%%%%%%%%%%%%%%%%%%%%%%%%%%%%%%%%%%%%%%%%%%%%%%%%%%%%%%%%%%%%%%%%%%%%%%%%%%%%
\section{Minimax defense against network-based attack} \label{sec:minimax-attnet}
%%%%%%%%%%%%%%%%%%%%%%%%%%%%%%%%%%%%%%%%%%%%%%%%%%%%%%%%%%%%%%%%%%%%%%%%%%%%%%%%

In this section, we consider another class of attacks -- the neural-network based attacks.
We present an algorithm for finding minimax solutions for this attack class,
and contrast the minimax solution with saddle-point and maximin solutions.

\vspace{-0.1in}
\subsection{Learning-based attacks}\label{sec:learning-based}

%An attacker ${z}(x):\mathcal{X} \to \mathcal{X}$ can be more general 
%than a specific class of attacks such as FGSM. 
Again, let $g:\mathcal{X} \to \mathcal{Y}$ is a classifier parameterized by $u$ 
and $l(g(x;u),y)$ is a loss function. 
The class of adversarial patterns $Z$ for the general minimax problem (Eq.~\ref{eq:minimax})
is very large, which results in strong but non-generalizable adversarial examples. 
Non-generalizable means the perturbation $z(x)$ has to be recomputed by solving the
optimization for every new test sample $x$. 
While such an ideal attack is powerful, its large size makes it
difficult to analytically study the optimal defense methods.
In Sec.~\ref{sec:minimax-grad}, we restricted this class to gradient-based attacks.
In this section, we restrict the class of patterns $Z$ to that which can be generated 
by a flexible but manageable class of perturbation $\{z(\cdot;v)\;|\;\forall v\in V\}$,
e.g., a neural network of a fixed architecture where the parameter $v$ is the network weights. 
This class is a clearly a subset of general attacks, but is generalizable, 
i.e., no time-consuming optimization is required in the test phase but only single 
feedforward passes though the network. 
The attack network ({\bf AttNet}), as we will call it, can be of any class of appropriate neural networks.
Here we use a three-layer fully-connected ReLU network with 300 hidden units per layer
in this paper.
Different from \cite{nguyen2017learning} or \cite{baluja2017adversarial}, 
we feed the label $y$ into the input of the network along with the features $x$.
This is analogous to using the true label $y$ in the original FGSM.
While this label input is not necessary but it can make the training of the attacker network
easier. 
As with other attacks, we impose the $l_\infty$-norm constraint on $z$, i.e., 
$\|z(x) - x \|_{\infty} \leq \eta$. 

Suppose now $f(u,v)$ is the empirical risk of a classifier-attacker pair
where the input $x$ is first transformed by attack network $z(x;v)$
and then fed to the classifier $g(z(x;v);u)$.
The attack network can be trained by gradient descent as well.
Given the classifier parameter $u$, we can use gradient descent
\begin{equation}
v \leftarrow v + \sigma \frac{\partial f(u,v)}{\partial v}
\end{equation}
to find an optimal attacker $v$ that {\it maximizes} the risk $f$ for the
given fixed classifier $u$.
Table~\ref{tbl:attnet_mnist} compares the error rates of the FGSM attacks
and the attack network (AttNet). 
The table shows that AttNet is better than or comparable to FGSM in all cases.
In particular, we already observed that the FGSM attack is not effective against
the classifier hardened against gradient-based attacks (Adv FGSM80 or Minimax-Grad),
but the AttNet can incur significant error ($>\sim 0.9$) for those hardened defenders for 
$\eta\geq0.2$. 
This indicates that the class of learning-based attacks is indeed different from 
the class of gradient-based attacks. 

\begin{table}[htb]
\small
%\centering
%\begin{center}
\renewcommand{\arraystretch}{0.9}
\centering
\begin{tabular}{|c||cc|c|cc|c|}
\hline
{Defense\textbackslash Attack} & FGSM-curr & AttNet-curr & worst & FGSM-curr & AttNet-curr & worst\\
\hline
& \multicolumn{3}{|c|}{$\eta$=0.1} & \multicolumn{3}{|c|}{$\eta$=0.2}\\
\cline{2-7}
No defense  & 0.446 & 0.697 & 0.697 & 0.933 & 0.999 & 0.999 \\  
Adv FGSM1   & 0.435 & 0.909 & 0.909 & 0.816 & 0.897 & 0.897 \\
Adv FGSM80  & 0.117 & 0.786 & 0.768 & 0.131 & 1.000 & 1.000 \\
Minimax-Grad& 0.025 & 0.498 & 0.498 & 0.085 & 0.956 & 0.956 \\
\hline
& \multicolumn{3}{|c|}{$\eta$=0.3} & \multicolumn{3}{|c|}{$\eta$=0.4}\\
\cline{2-7}
No defense  & 0.983 & 1.000 & 1.000 & 0.985 & 1.000 & 1.000 \\  
Adv FGSM1   & 0.892 & 1.000 & 1.000 & 0.898 & 1.000 & 1.000 \\
Adv FGSM80  & 0.352 & 0.887 & 0.887 & 0.688 & 1.000 & 1.000 \\
Minimax-Grad& 0.144 & 1.000 & 1.000 & 0.221 & 1.000 & 1.000 \\
\hline
%& \multicolumn{3}{|c|}{$\eta$=0.2} & \multicolumn{3}{|c|}{$\eta$=0.3}\\
%\cline{2-7}
%No defense & 0.933 & 0.999 & 0.999 & 0.983 & 1.000 & 1.000\\  
%Adv FGSM1  & 0.816 & 0.897 & 0.897 & 0.892 & 1.000 & 1.000\\
%Adv FGSM80 & 0.131 & 1.000 & 1.000 & 0.043 & 0.887 & 0.887 \\
%Minimax-Grad  & 0.027 & 0.956 & 0.956 & 0.036 & 1.000 & 1.000\\
%\hline
%& \multicolumn{3}{|c|}{$\eta$=0.4} & \multicolumn{3}{|c|}{$\eta$=0.5}\\
%\cline{2-7}
%No defense & 0.985 & 1.000 & 1.000 & 0.987 & 1.000 & 1.000\\  
%Adv FGSM1  & 0.898 & 1.000 & 1.000 & 0.859 & 0.902 & 0.902\\
%Adv FGSM80 & 0.031 & 1.000 & 1.000 & 0.066 & 0.899 & 0.899\\
%Minimax-Grad  & 0.026 & 1.000 & 1.000 & 0.016 & 0.903 & 0.903\\
%\hline
\end{tabular} 
\caption{Test error rates of FGSM vs learning-based attack network (AttNet)
on various adversarially-trained classifiers for MNIST. FGSM-curr/AttNet-curr means
they are computed/trained for the specific classifier on the left.
Worst means the larger of FGSM-curr and AttNet errors for each $\eta$. 
Note that FGSM fails to attack hardened networks (Adv FGSM80 and Minimax-Grad), 
whereas AttNet can still attack them successfully.}
\label{tbl:attnet_mnist}
%\end{center}
\end{table}

\vspace{-0.5in}

\subsection{Minimax solution}
%\subsection{Minimax and saddle-point solutions}

We consider the two-player zero-sum game between a classifier and an network-based
attacker where each player can choose its own parameters.
Given the current classifier $u$, an optimal whitebox attacker parameter $v$
is the maximizer of the risk $f(u,v)$
\begin{equation}\label{eq:vs(s)}
\vs(u)\triangleq \arg\max_v f(u,v).
\end{equation}
Consequently, the defender should choose the classifier parameters $u$ 
such that the maximum risk is minimized
\begin{equation}\label{eq:us}
\us \triangleq \arg\min_u \max_v f(u,v) = \arg\min_u f(u,\vs(u)). 
\end{equation}
%where $\vs(u)$ is the maximizer function
%\begin{equation}\label{eq:vs(s)}
%\vs(u) \triangleq \arg\max_v f(u,v).
%\end{equation}
As before, this solution to the continuous minimax problem
has a natural interpretation as the best worst-case solution.
Assuming the attacker is optimal, i.e., it chooses the best attack 
from Eq.~\ref{eq:vs(s)} given $u$, no other defense can achieve a lower risk than
the minimax defense $\us$ in Eq.~\ref{eq:us}. 
The minimax defense is also a conservative defense. If the attacker is not optimal, 
and/or if the attack does not know the defense $u$ exactly (as in blackbox attacks), 
the actual risk can be lower than what the minimax solution $f(\us,\vs(\us))$ predicts. 
Before proceeding further, we point out that the claims above apply to the global minimizer
$\us$ and the maximizer function $\vs(\cdot)$, 
but in practice we can only find local solutions for complex risk functions of 
deep classifiers and attackers.

To solve Eq.~\ref{eq:us}, we analyze the problem similarly to 
Eqs.~\ref{eq:gd1}-\ref{eq:sensitivity} from the previous section.
At each iteration, the defender should choose $u$ in expectation of the attack 
and minimize $f(u,\vs(u))$. 
We use gradient descent 
\begin{equation}
u \leftarrow u - \lambda \frac{df(u,\vs(u))}{du},
\end{equation}
where the total derivative $\frac{df}{du}$ is
\begin{equation}\label{eq:gd2}
\frac{df}{du} = \frac{\partial f(u,v^\ast(u))}{\partial u} + \frac{\partial v^\ast(u)}{\partial u} \frac{\partial f(u,v)}{\partial v}.
\end{equation}
Since the exact maximizer $v^\ast(u)$ is difficult to find, we only update $v$
incrementally by one (or more) steps of gradient-ascent update
\begin{equation}\label{eq:gd3}
v \leftarrow v + \sigma \frac{\partial f(u,v)}{\partial v}. 
\end{equation}
The resulting formulation is closely related to the unrolled optimization \citep{metz2016unrolled} proposed for training
GANs, although the latter has a very different cost function $f$.
Using the single update (Eq.~\ref{eq:gd3}), the total derivative is
\begin{equation}
\frac{df}{du} = \frac{\partial f(u,v^\ast(u))}{\partial u} + \sigma \frac{\partial^2 f(u,v)}{\partial u \partial v} \frac{\partial f(u,v)}{\partial v}.
\end{equation}
Similar to hardening a classifier against gradient-based attacks by minimizing 
Eq.~\ref{eq:sensitivity} at each iteration, 
the gradient update of $u$ for $f(u,v)$ can be done using the gradient of the
following sensitivity-penalized function
\begin{equation}
f_{\mathrm{sens}}(u) \triangleq f(u,v) + \frac{\sigma}{2} \left\|\frac{\partial f(u,v)}{\partial v}\right\|^2.
\end{equation}
In other words, $u$ is chosen not only to minimize the risk but also to prevent
the attacker from exploiting the sensitivity of $f$ to $v$. 
The algorithm is summarized in Alg.~\ref{alg:minimax}. 

\begin{algorithm}[htb] \caption{Minimax Optimization by Sensitivity Penalization} \label{alg:minimax}
{Input}: risk $f(u,v)$, \# of iterations $T$, learning rates $(\sigma_i)$, $(\lambda_{i})$, $(\gamma_i)$\\
{Output}: $(\us, \vs(\us))$ \\
Initialize $u_0, v_0$\\
{Begin}
\begin{algorithmic}
\FOR{$i=1,\;...\;,T$}
	\STATE\hspace{-0.075in}{{\it Max step:} $v_i = v_{i-1} + \sigma_i\;\frac{\partial f(u_{i-1},v_{i-1})}{\partial v}$}
	\STATE\hspace{-0.075in}{{\it Min step:}  $u_i = u_{i-1} - \lambda_i \frac{\partial}{\partial u} \left[f(u_{i-1},v_{i-1})+\frac{\gamma_{i-1}}{2}\left\|\frac{\partial f(u_{i-1},v_{i-1})}{\partial v}\right\|^2\right]$.}
\ENDFOR
\STATE{Return ($u_T$, $v_T$).}
\end{algorithmic}
%\vspace{0.1in}
\end{algorithm}
The classifier obtained after convergence will be referred to as {\bf Minimax-AttNet}. 
Note that Alg.~\ref{alg:minimax} is independent of the adversarial example problems
presented in the paper, and can be used for other minimax problems as well. 

\subsection{Minimax vs maximin solutions}

In analogy with the minimax problem, we can also consider the maximin solution defined by
\begin{equation}\label{eq:vs}
\vs \triangleq \arg\max_v \min_u f(u,v) = \arg\max_v f(\us(v),v).
\end{equation}
where
\begin{equation}\label{eq:us(v)}
\us(v) \triangleq \arg\min_u f(u,v)
\end{equation}
is the minimizer function. 
Here we are abusing the notations for the minimax solution $\us$,
the maximin solution $\vs$, the minimizer $\us(\cdot)$, and the maximizer $\vs(\cdot)$. 
%Changing the role of leader/follower often makes no sense.
%Think of GAN: find a classifier that can discriminate the outputs of even 
%the best generator? Doesn't make sense. 
%But finding a generator that can fool even the best discriminator does. 
Similar to the minimax solution, the maximin solution has an intuitive meaning --
it is the best worst-case solution for the attacker.
Assuming the defender is optimal, i.e., it chooses the best defense from Eq.~\ref{eq:us(v)}
that minimizes the risk $f(u,v)$ given the attack $v$,
no other attack can inflict a higher risk than the maximin attack $\vs$. 
It is also a conservative attack. If the defender is not optimal, 
and/or if the defender does not know the attack $v$ exactly, 
the actual risk can be higher than what the solution $f(\us(\vs),\vs)$ predicts. 
Note that the maximin scenario where the defender knows the attack method is not very
realistic but is the opposite of the minimax scenario and provides the lower bound.

To summarize, minimax and maximin defenses and attacks have the following inherent properties.
\begin{lemma}\label{thm:properties}
Let $\us, \vs(u), \vs, \us(v)$ be the solutions of Eqs.~\ref{eq:us},\ref{eq:vs(s)},\ref{eq:vs},\ref{eq:us(v)}.
\begin{enumerate}
\item $f(u,\vs(u)) \geq f(u,v)$: For any given defense $u$, the max attack $\vs(u)$
is the most effective attack.
\item $f(\us,\vs(\us)) \leq f(u,\vs(u))$: Against the optimal attack $\vs(u)$, 
the minimax defense $\us$ is the most effective defense.
\item $f(\us(v),v) \leq f(u,v)$: For any given attack $v$, the min defense $\us(v)$
is the most effective defense. 
\item $f(\us(v),\vs) \geq f(\us(v),v)$: Against the optimal defense $\us(v)$, 
the maximin attack $\vs$ is the most effective attack.
\item $\max_v \min_u f(u,v) \leq \min_u \max_v f(u,v)$: 
The risk of the best worst-case attack is lower than
that of the best worst-case defense. 
\end{enumerate}
\end{lemma}
These properties follow directly from the definitions. 
The lemma helps us to better understand the dependence of defense and attack,
and gives us the range of the possible risk values which can be measured empirically.
\if0
Note that it does not mean $\max_v \min_u f(u,v) \leq f(u,v) \leq \min_u \max_v f(u,v)$
for arbitrary $(u,v)$.  These bounds are for optimal attacks and /defenses with the knowledge of the opponent.
The full range of $f(u,v)$ is given by
$\min_u \min_v f(u,v) \leq f(u,v) \leq \max_u \max_v f(u,v)$, which is not very informative.
\fi
To find maximin solutions, we use the same algorithm (Alg.~\ref{alg:minimax})
except that the variables $u$ and $v$ are switched and the sign of $f$ is flipped
before the algorithm is called. 
The resultant classifier will be referred to as Maximin-AttNet.

\subsection{Experiments}\label{sec:experiments2}

\begin{figure*}[thb]
\centering
%This is a placeholder.
\includegraphics[width=1\linewidth]{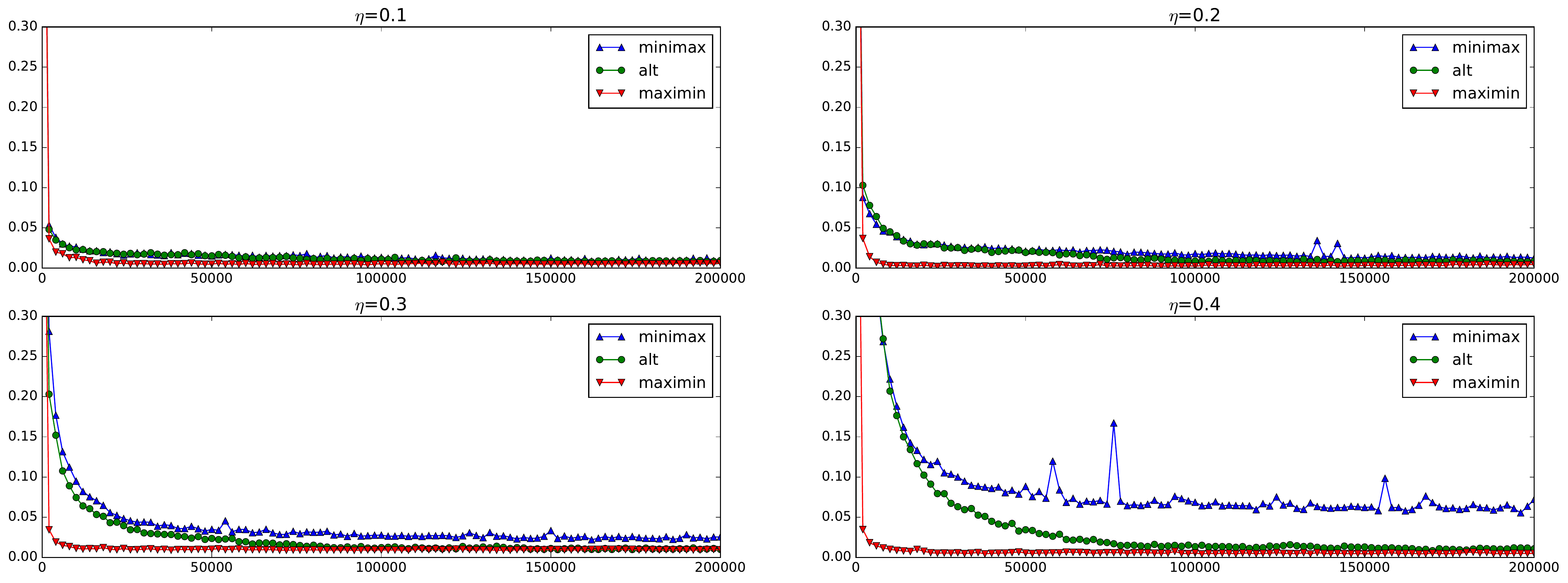}
\caption{Convergence of the test error rates for Minimax-AttNet (blue),
Alt-AttNet (green), and Maximin-AttNet (red) for MNIST.
}
\label{fig:minimax_mnist}
\end{figure*}

\vspace{-0.1in}
In addition to minimax and maximin optimization, we also consider as a reference algorithm
the {alternating descent/ascent} method used in GAN training \cite{goodfellow2014generative}
\begin{equation}
u \leftarrow u - \lambda \frac{\partial f}{\partial u},\;\;\;\;v \leftarrow v + \sigma \frac{\partial f}{\partial v},
\end{equation}
and refer to its solution as Alt-AttNet.
Similar to our discussion on Minimax-Grad and LWA in Sec.~\ref{sec:experiments1},
the alternating descent/ascent finds local saddle points
which are not necessarily minimax or maximin solutions, and therefore its solution
will in general be different from the solution from Alg.~\ref{alg:minimax}.
The difference of the solutions from three optimizations -- {Minimax-AttNet, Maximin-AttNet, 
and Alt-AttNet} -- applied to a common problem, is demonstrated in 
Fig.~\ref{fig:minimax_mnist}. The figure shows the test error over the course of
optimization starting from random initializations. 
One can see that Minimax-AttNet (top blue curves) and Alt-AttNet (middle green curves) 
converge to different values suggesting the learned classifiers will also be different.

Table~\ref{tbl:minimax_vs_others_mnist} compares the robustness of 
the classifiers -- Minimax-AttNet, Alt-AttNet and Minimax-Grad (from Sec.~\ref{sec:minimax-grad}),
against the AttNet attack. and the FGSM attack.
Not surprisingly, both Minimax-AttNet and Alt-AttNet are much more robust than Minimax-Grad
against AttNet. 
Minimax-AttNet performs similarly to Alt-AttNet at $\eta$=0.1 and 0.2, but is much better
at $\eta$=0.3 and 0.4. 
The different performance of Minimax-AttNet vs Alt-AttNet implies that the minimax solution
found by Alg.~\ref{alg:minimax} is different from the solution found by alternating
descent/ascent. 
In addition, against FGSM attacks, Minimax-AttNet is quite robust (0.058 - 0.116)
despite that the classifiers are not trained against gradient-based attacks at all. 
In contrast, Minimax-Grad is very vulnerable (0.498 -- 1.000) against AttNet
which we have already observed. 
This result suggests that the class of AttNet attacks and the class of gradient-based
attacks are indeed different, and the former class partially subsumes the latter. 

\begin{table}[htb]
%\centering
\small
%\begin{center}
\renewcommand{\arraystretch}{0.9}
\centering
\begin{tabular}{|c||cc|c|cc|c|}
\hline
{Defense\textbackslash Attack} & FGSM-curr & AttNet-curr & worst & FGSM-curr & AttNet-curr & worst\\
\hline
& \multicolumn{3}{|c|}{$\eta$=0.1} & \multicolumn{3}{|c|}{$\eta$=0.2} \\
\cline{2-7}
Minimax-AttNet & 0.058 & 0.010 & 0.058 & 0.109 & 0.010 & 0.109 \\
Alt-AttNet     & 0.048 & 0.010 & 0.048 & 0.096 & 0.016 & 0.096 \\
Minimax-Grad   & 0.025 & 0.498 & 0.498 & 0.085 & 0.956 & 0.956 \\
\hline
& \multicolumn{3}{|c|}{$\eta$=0.3} & \multicolumn{3}{|c|}{$\eta$=0.4} \\
\cline{2-7}
Minimax-AttNet & 0.116 & 0.018 & 0.116 & 0.079 & 0.364 & 0.364 \\
Alt-AttNet     & 0.158 & 0.032 & 0.158 & 0.334 & 0.897 & 0.897 \\
Minimax-Grad   & 0.144 & 1.000 & 1.000 & 0.221 & 1.000 & 1.000 \\
\hline
%& \multicolumn{3}{|c|}{$\eta$=0.2} & \multicolumn{3}{|c|}{$\eta$=0.3} \\
%\cline{2-7}
%Minimax-AttNet & 0.109 & 0.010 & 0.109 & 0.116 & 0.018 & 0.116\\
%Alt-AttNet     & 0.096 & 0.016 & 0.096 & 0.158 & 0.032 & 0.158\\
%Minimax-Grad   & 0.027 & 0.956 & 0.956 & 0.036 & 1.000 & 1.000\\
%\hline
%& \multicolumn{3}{|c|}{$\eta$=0.4} & \multicolumn{3}{|c|}{$\eta$=0.5} \\
%\cline{2-7}
%Minimax-AttNet & 0.079 & 0.364 & 0.364 & 0.337 & 1.000 & 1.000\\
%Alt-AttNet     & 0.334 & 0.897 & 0.897 & 0.347 & 1.000 & 1.000\\
%Minimax-Grad   & 0.026 & 1.000 & 1.000 & 0.016 & 0.903 & 0.903\\
%\hline
\end{tabular} 
\caption{Test error rates of Minimax-AttNet, Alt-AttNet, and Minimax-Grad
classifiers for MNIST. 
Worst means the larger of FGSM-curr and AttNet errors for each row. 
Minimax-AttNet is better than Alt-AttNet and Minimax-Grad overall, 
and is also moderately robust against the out-of-class attack (FGSM-curr).
}
\label{tbl:minimax_vs_others_mnist}
%\end{center}
\end{table}

Lastly, adversarial examples generated by various attacks in the paper have
diverse patterns and are shown in Fig.~\ref{fig:adversarial_examples}
of the appendix.
All the experiments with the MNIST dataset presented so far are also performed
with the CIFAR-10 dataset and are reported in the appendix.
To summarize, the results with CIFAR-10 are similar: Minimax-Grad outperforms non-minimax defenses, 
and AttNet can attack classifiers which are hardened against gradient-based attacks. 
However, gradient-based attacks are also very effective against classifiers 
hardened against AttNet, and also Minimax-AttNet and Alt-AttNet perform similarly,
which are not the case with the MNIST dataset. 
The issue of defending against out-of-class attacks is discussed in the next section.

%\vspace{-0.1in}
%%%%%%%%%%%%%%%%%%%%%%%%%%%%%%%%%%%%%%%%%%%%%%%%%%%%%%%%%%%%%%%%%%%%%%%%%%%%%%%%
\section{Discussion}\label{sec:discussion}
%%%%%%%%%%%%%%%%%%%%%%%%%%%%%%%%%%%%%%%%%%%%%%%%%%%%%%%%%%%%%%%%%%%%%%%%%%%%%%%%

\vspace{-0.1in}
\subsection{Robustness against multiple attack types}
%\vspace{-0.1in}
We discuss some limitations of the current study and propose an extension. 
Ideally, a defender should find a robust classifier against the worst attack
from a large class of attacks such as the general minimax problem (Eq.~\ref{eq:minimax}). 
However, it is difficult to train classifiers with a large class of attacks, due to
the difficulty of modeling the class and of optimization itself. 
% and the class has to be smaller such as the class of neural networks used in this paper. 
On the other hand, if the class is too small, then the worst attack from that class
is not representative of all possible worst attacks, and therefore the minimax defense 
found will not be robust to out-of-class attacks. The trade-off seems inevitable. 
\if0
Also, the three-layer attack network used in the paper outperforms FGSM for MNIST data
but not always for CIFAR-10 data (Table~\ref{tbl:attnet_cifar10} in the appendix),
likely due to the increased complexity of training. It remains to clarify the cause
with more experiments.
\fi
%To summarize, an ideal defense robust against all types of attack is still elusive,
%but minimax defense using AttNet is a much better choice than 
%the defense hardened against only specific types of attacks such as FGSM. 

%As we saw in Sec~\ref{sec:experiments}, minimax defense against AttNet is 
%still moderately vulnerable to FGSM attaks. 

It is, however, possible to build a defense against multiple specific
types of attacks. 
Suppose $z_1(u),...,z_m(u)$ are $m$ different types of attacks, e.g., 
$z_1$=FGSM, $z_2$=IFGSM, etc.
The minimax defense for the combined attack is the solution to 
the mixed continuous-discrete problem
\begin{equation}
\min_u \max \{f(u,z_1(u)), ... ,f(u,z_m(u))\}.
\end{equation}
Additionally, suppose $z_{m+1}(u,v),...,z_{m+n}(u,v)$ are $n$ different types of
learning-based attacks, e.g., $z_{m+1}$=2-layer dense net, $z_{m+2}$=5-layer convolutional
nets, etc. 
The minimax defense against the mixture of multiple fixed-type and learning-based attacks
can be found by solving 
\begin{equation}\label{eq:mixture}
\min_u \max \{f(u,z_1(u)), ...\; ,f(u,z_m(u)), \;\max_v f(u,z_{m+1}(u,v)), ... \;, \max_v f(u,z_{m+n}(u,v)) \},
\end{equation}
that is, minimize the risk against the strongest attacker across multiple attack classes. 
Note the strongest attacker class and its parameters change as the classifier $u$ changes. 
Due to the computational demand to solve Eq.~\ref{eq:mixture}, we leave it as 
a future work to compute minimax solutions against multiple classes of attacks.

\subsection{Adversarial examples and privacy attacks}

Lastly, we discuss a bigger picture of the game between adversarial players. 
The minimax optimization arises in the leader-follower game \citep{bruckner2011stackelberg}
with the constant sum constraint. The leader-follower setting makes sense because the
defense (=classifier parameters) is often public knowledge and the attacker
exploits the knowledge. 
Interestingly, the problem of the attack on privacy \citep{hamm2017minimax} 
has a very similar formulation as the adversarial attack problem, 
different only in that the classifier is an attacker and the data perturbator is a defender.
In the problem of privacy preservation against inference, the defender is
a data transformer $z(x)$ (parameterized by $u$) which perturbs the raw data,
and the attacker is a classifier (parameterized by $v$) who tries to extract sensitive
information such as identity from the perturbed data such as online activity of a person. 
The transformer is the leader, such as when the privacy mechanism is public knowledge, 
and the classifier is the follower as it attacks the given perturbed data. 
The risk for the defender is therefore the accuracy of the inference of
sensitive information measured by $-E[l(z(x;u),y;v)]$. 
%Since a optimal attacker (=classifier) tries to maximize the accuracy of the inference of
%sensitive information ($\max_v -E[l(z(x;u),y;v)]$), the defender (=transformer) should
%should transform the raw data to minimize this maximum accuracy. 
Solving the minimax risk problem ($\min_u \max_v -E[l(z(x;u),y;v)]$) gives us 
the best worst-case defense when the classifier/attacker knows the transformer/defender
parameters, which therefore gives us a robust data transformer to preserve the privacy
against the best inference attack (among the given class of attacks.)
On the other hand, solving the maximin risk problem ($\max_v \min_u -E[l(z(x;u),y;v)]$)
gives us the best worst-case classifier/attacker when its parameters are known to the
transformer. 
As one can see, the problems of adversarial attack and privacy attack are two sides of 
the same coin, which can potentially be addressed by similar frameworks and optimization algorithms.

%%%%%%%%%%%%%%%%%%%%%%%%%%%%%%%%%%%%%%%%%%%%%%%%%%%%%%%%%%%%%%%%%%%%%%%%%%%%%%%%
\section{Conclusion} \label{sec:conclusion}
%%%%%%%%%%%%%%%%%%%%%%%%%%%%%%%%%%%%%%%%%%%%%%%%%%%%%%%%%%%%%%%%%%%%%%%%%%%%%%%%

In this paper, we explain and formulate the adversarial sample problem in the context of
two-player continuous game. 
We analytically and numerically study the problem with two types of attack classes
-- gradient-based and network-based -- and show different properties of the solutions
from those two classes. 
While a classifier robust to all types of attack may yet be an elusive goal, 
we claim that the minimax defense is a very reasonable goal, and that such a defense
can be computed for classes such as gradient- or network-based attacks.
We present optimization algorithms for numerically finding those defenses. 
The results with the MNIST and the CIFAR-10 dataset show that the classifier 
found by the proposed method outperforms non-minimax optimal classifiers, 
and that the network-based attack
is a strong class of attacks that should be considered in adversarial example research 
in addition to the gradient-based attacks which are used more frequently.
As future work, we plan to study further on the issue of transferability of a defense
method to out-of-class attacks, and on efficient minimax optimization algorithms for 
finding defense against general attacks.
%Due to the complex landscape of loss functions and the added difficulty of multilevel
%optimization, the local solution found can be suboptimal and warrants further study
%on efficient minimax optimization algorithms.
%We show that the proposed optimization method can find minimax defenses 
%which are more robust than adversarially-trained classifiers and the classifiers
%from simple alternating descent/ascent.
%We demonstrate these with MNIST and CIFAR-10. 
%In this paper, we demonstrated with the MNIST dataset, it is easy to attack a fixed defense and 
%also easy to defend against a fixed attack. 
%However, the problem becomes more involved for both if one player chooses her action 
%in expectation of the other player.
%The paper not only provides conceptual tool to model the interaction of attacks
%and defense, but also practical algorithms that can find feasible attacks and defenses,
%which are theoretically the best worst-case options given the class of attack/defense
%models.
%Using MNIST and CIFAR-10 datasets, we show that the learned minimax and the maximin models
%using simple neural networks indeed displays the behavior of the best worst case
%models. 

%\clearpage
%\pagebreak

%\section*{Acknowledgements} 

{\small
\bibliography{icml18_jh,gamesec18_jh}
}

%\clearpage
%\pagebreak

\appendix

\section{Results with MNIST}

%\subsection{Network structure and training}

The architecture of the MNIST classifier is similar to the Tensorflow model\footnote{\url{https://github.com/tensorflow/models/tree/master/tutorials/image/mnist}},
and is trained with the following hyperparameters:\\
$\{${\it Batch size = 128, optimizer = AdamOptimizer with $\lambda=10^{-4}$, total \# of iterations=50,000.}$\}$ 

The attack network has three hidden fully-connected layers of 300 units, 
trained with the following hyperparameters:\\
$\{${\it Batch size = 128, dropout rate = 0.5, optimizer = AdamOptimizer with $10^{-3}$, total \# of iterations=30,000.}$\}$ 

For minimax, saddle-point, and maximin optimization, the total number of iteration was
100,000. 
The sensitivity-penalty coefficient of $\gamma=1$ was used in Alg.~\ref{alg:minimax}.

\begin{figure*}[thb]
\centering
\includegraphics[width=1\linewidth]{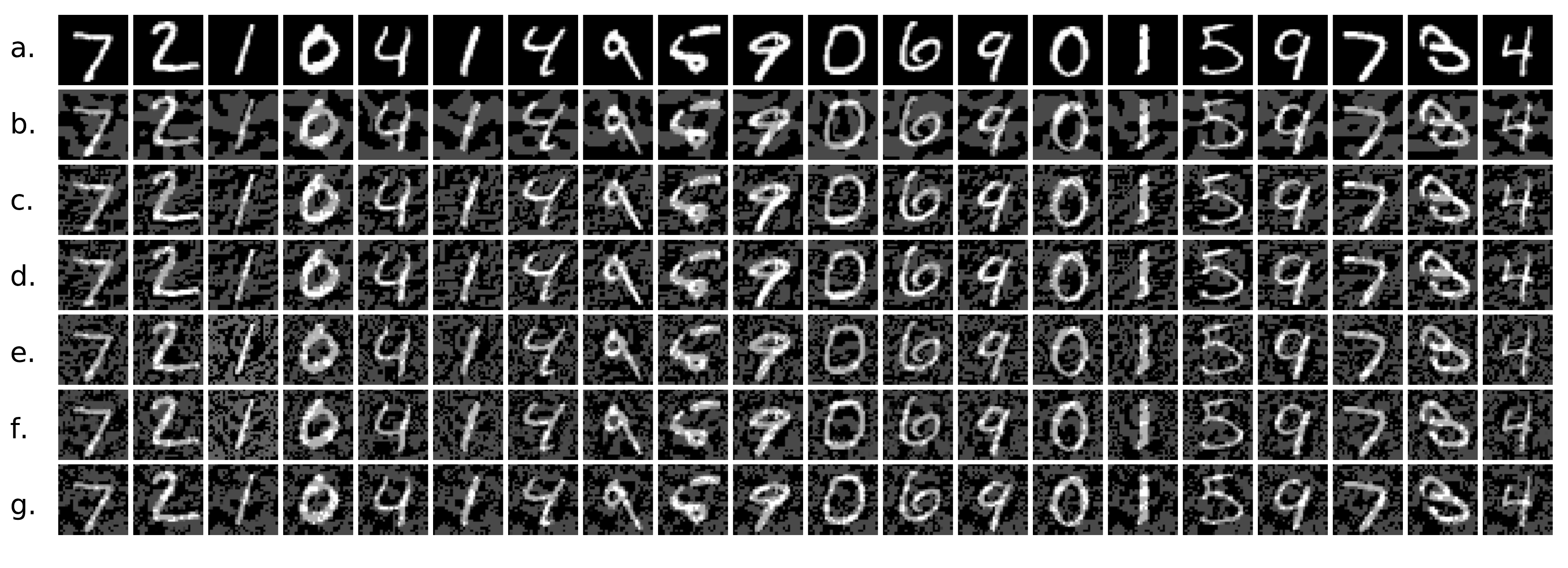}
\caption{Adversarial samples generated from different attacks at $\eta=0.2$. 
(a) Original data  (b) FGSM1 (c) FGSM80 (d) IFGSM1 (e) Minimax-AttNet (f) Alt-AttNet (g) Maximin-AttNet. Note the diversity of patterns.
}
\label{fig:adversarial_examples}
\end{figure*}

\section{Results with CIFAR-10}

We preprocess the CIFAR-10 dataset by removing the mean and normalizing the pixel
values with the standard deviation of all pixels in the image. 
It is followed by clipping the values to $\pm 2$ standard deviations
and rescaling to $[-1,1]$.
The architecture of the CIFAR classifier is similar to the Tensorflow model\footnote{\url{
https://github.com/tensorflow/models/tree/master/tutorials/image/cifar10}}
but is simplified further by removing the local response normalization layers.
With the simple structure, we attained $\sim78\%$ accuracy with the test data.
The classifier is trained with the following hyperparameters:\\
$\{${\it Batch size = 128, optimizer = AdamOptimizer with $\lambda=10^{-4}$, total \# of iterations=100,000.}$\}$

The attack network has three hidden fully-connected layers of 300 units, 
trained with the following hyperparameters:\\
$\{${\it Batch size = 128, dropout rate = 0.5, optimizer = AdamOptimizer with $\sigma=10^{-3}$, total \# of iterations=30,000.}$\}$

For minimax, saddle-point, and maximin optimization, the total number of iteration was
100,000. The sensitivity-penalty coefficient of $\gamma=1$ was used in Alg.~\ref{alg:minimax}.

In the rest of the appendix, we repeat all the experiments with the MNIST dataset 
using the CIFAR-10 dataset.

\begin{table}[htb]
\renewcommand{\arraystretch}{0.9}
\small
\centering
%\begin{center}
\setlength\tabcolsep{4pt}
\begin{tabular}{|c||c|cccc|}
\hline
\multirow{2}{*}{Defense\textbackslash Attack} & \multirow{2}{*}{No attack} & 
\multicolumn{4}{c|}{FGSM} \\
\cline{3-6}
% & & $\eta$=0.01 & $\eta$=0.02 & $\eta$=0.03 & $\eta$=0.04 \\
 & & $\eta$=0.05 & $\eta$=0.06 & $\eta$=0.07 & $\eta$=0.08 \\ 
\hline 
%No defense & 0.222 & 0.588 & 0.649 & 0.679 & 0.738 \\
No defense & 0.222 & 0.766 & 0.790 & 0.807 & 0.823\\
\hline
\end{tabular} 
\caption{
Test error rates of the FGSM attack on an undefended convolutional neural network for CIFAR-10.
Higher error means a more successful attack.}
\label{tbl:known_defense_cifar10}
%\end{center}
\end{table}

\begin{table}[htb]
\renewcommand{\arraystretch}{0.9}
\small
%\begin{center}
\centering
\setlength\tabcolsep{4pt}
\begin{tabular}{|c||c|cccc|}
\hline
\multirow{2}{*}{Defense\textbackslash Attack} & \multirow{2}{*}{No attack} & 
\multicolumn{4}{c|}{FGSM} \\
\cline{3-6}
% & & $\eta$=0.01 & $\eta$=0.02 & $\eta$=0.03 & $\eta$=0.04 \\
 & & $\eta$=0.05 & $\eta$=0.06 & $\eta$=0.07 & $\eta$=0.08 \\
\hline
%Adv train & n/a & 0.330 & 0.364 & 0.383 & 0.408 \\
Adv train & n/a & 0.425 & 0.452 & 0.466 & 0.470 \\
\hline
\end{tabular} 
\caption{Test error rates of the FGSM attacks on adversarially-trained classifiers for CIFAR-10. 
This defense can significantly lower the errors from the attacks, although not
as low as the MNIST case.}
\label{tbl:known_attack_cifar10}
%\end{center}
\end{table}

\begin{table}[htb]
\renewcommand{\arraystretch}{0.9}
\small
%\begin{center}
\centering
\begin{tabular}{|c|c||c|cccc|c|}
\cline{2-8}
\multicolumn{1}{c|}{}
&\multirow{2}{*}{Defense\textbackslash Attack} & \multirow{2}{*}{No attack} & 
\multicolumn{4}{|c|}{FGSM} & \multirow{2}{*}{FGSM-curr} \\
\cline{4-7}
\multicolumn{1}{c|}{}& & &FGSM-1 & FGSM-2 & $\cdots$ & FGSM-40 &  \\
\cline{2-8} \hline

\multirow{5}{*}{$\eta$=0.05}
&No defense  & 0.222 & 0.766 & 0.734 & $\cdots$ & 0.655 & 0.766  \\
&Adv FGSM1  & 0.215 & 0.425 & 0.533 & $\cdots$ & 0.420 & 0.533  \\
&Adv FGSM2  & 0.206 & 0.422 & 0.456 & $\cdots$ & 0.406 & 0.501  \\
&Adv FGSM40  & 0.210 & 0.370 & 0.412 & $\cdots$ & 0.348 & 0.588  \\
&LWA        & 0.203 & 0.422 & 0.464 & $\cdots$ & 0.423 & 0.456  \\
&Minimax-Grad  & 0.203 & 0.425 & 0.475 & $\cdots$ & 0.423 & 0.481 \\ 
\hline
\multirow{5}{*}{$\eta$=0.06}
&No defense  & 0.222 & 0.790 & 0.761 & $\cdots$ & 0.680 &  0.790  \\
&Adv FGSM1  & 0.215 & 0.452 & 0.565 & $\cdots$ & 0.440 & 0.565  \\
&Adv FGSM2  & 0.208 & 0.447 & 0.482 & $\cdots$ & 0.431 & 0.517  \\
&Adv FGSM40  & 0.216 & 0.398 & 0.431 & $\cdots$ & 0.353 & 0.599  \\
&LWA        & 0.208 & 0.446 & 0.493 & $\cdots$ & 0.447 &  0.489  \\
&Minimax-Grad  & 0.199 & 0.431 & 0.473 & $\cdots$ & 0.446  & 0.453 \\ 
\hline
\multirow{5}{*}{$\eta$=0.07}
& No defense  & 0.222 & 0.807 & 0.787 & $\cdots$ & 0.704 & 0.807  \\
&Adv FGSM1  & 0.214 & 0.466 & 0.555 & $\cdots$ & 0.450 &  0.555  \\
&Adv FGSM2  & 0.206 & 0.456 & 0.490 & $\cdots$ & 0.445 &  0.501  \\
&Adv FGSM40  & 0.218 & 0.397 & 0.416 & $\cdots$ & 0.346 & 0.423  \\
&LWA        & 0.208 & 0.453 & 0.499 & $\cdots$ & 0.451 & 0.485 \\
&Minimax-Grad  & 0.208 & 0.461 & 0.497& $\cdots$ & 0.456  & 0.487  \\
\hline
\multirow{5}{*}{$\eta$=0.08}
&No defense  & 0.222 & 0.823 & 0.807 & $\cdots$ & 0.709 & 0.823  \\
&Adv FGSM1  & 0.213 & 0.470 & 0.533 & $\cdots$ & 0.462 & 0.533  \\
&Adv FGSM2  & 0.204 & 0.459 & 0.466 & $\cdots$ & 0.462  & 0.476  \\
&Adv FGSM40  & 0.226 & 0.422 & 0.421 & $\cdots$ & 0.331 & 0.338  \\
&LWA        & 0.208 & 0.470 & 0.485 & $\cdots$ & 0.469 & 0.485  \\
&Minimax-Grad  & 0.203 & 0.456 & 0.464 & $\cdots$ & 0.459 & 0.462 \\ 
\hline
\end{tabular} 
\caption{Test error rates of different attacks on various adversarially-trained classifiers for CIFAR-10.
FGSM-curr means the FGSM attack on the specific classifier on the left.
Worst means the largest error in each row. 
Adv FGSM is the classifier adversarially trained with FGSM attacks.
Minimax-Grad is the result of minimizing Eq.~\ref{eq:sensitivity} by gradient descent.
LWA is the result of minimizing Eq.~\ref{eq:sensitivity} without the gradient-norm term.
}
\label{tbl:normtrain_cifar10}
%\end{center}
\end{table}

\begin{table}[htb]
\renewcommand{\arraystretch}{0.9}
\small
\centering
%\begin{center}
\begin{tabular}{|c||cc|c|cc|c|}
\hline
{Defense\textbackslash Attack} & FGSM-curr & AttNet-curr & worst & FGSM-curr & AttNet-curr & worst\\
\hline
\if0
& \multicolumn{3}{|c|}{$\eta$=0.01} & \multicolumn{3}{|c|}{$\eta$=0.02}\\
\cline{2-7}
No defense  & 0.588 & 0.238 & 0.588 & 0.649 & 0.278 & 0.649 \\  
Adv FGSM1   & 0.358 & 0.226 & 0.358 & 0.415 & 0.247 & 0.415 \\  
Adv FGSM40  & 0.323 & 0.217 & 0.323 & 0.375 & 0.247 & 0.375 \\  
Minimax-Grad& 0.342 & 0.212 & 0.342 & 0.390 & 0.242 & 0.390 \\  
\hline
 & \multicolumn{3}{|c|}{$\eta$=0.03} & \multicolumn{3}{|c|}{$\eta$=0.04}\\
\cline{2-7}
No defense  & 0.697 & 0.356 & 0.697 & 0.738 & 0.450 & 0.738 \\  
Adv FGSM1   & 0.464 & 0.276 & 0.464 & 0.505 & 0.305 & 0.505 \\  
Adv FGSM40  & 0.538 & 0.338 & 0.538 & 0.573 & 0.388 & 0.267 \\  
Minimax-Grad& 0.439 & 0.275 & 0.439 & 0.462 & 0.326 & 0.462 \\  
\hline
\fi
& \multicolumn{3}{|c|}{$\eta$=0.05} & \multicolumn{3}{|c|}{$\eta$=0.06}\\
\cline{2-7}
  no defense  & 0.766 & 0.504 & 0.766 & 0.583 & 0.766 & 0.790 \\
   Adv FGSM1  & 0.533 & 0.356 & 0.533 & 0.565 & 0.473 & 0.565 \\
  Adv FGSM40  & 0.588 & 0.454 & 0.588 & 0.599 & 0.442 & 0.599 \\
Minimax-Grad  & 0.481 & 0.343 & 0.481 & 0.453 & 0.484 & 0.484 \\
\hline
& \multicolumn{3}{|c|}{$\eta$=0.07} & \multicolumn{3}{|c|}{$\eta$=0.08}\\
\cline{2-7}
  no defense  & 0.807 & 0.655 & 0.807 & 0.823 & 0.685 & 0.823 \\
   Adv FGSM1  & 0.555 & 0.499 & 0.555 & 0.535 & 0.678 & 0.678 \\
  Adv FGSM40  & 0.423 & 0.669 & 0.669 & 0.338 & 0.797 & 0.797 \\
Minimax-Grad  & 0.487 & 0.529 & 0.529 & 0.462 & 0.607 & 0.607 \\
\hline
\end{tabular} 
\caption{Test error rates of FGSM vs learning-based attack network (AttNet)
on various adversarially-trained classifiers for CIFAR-10. FGSM-curr/AttNet-curr means
they are computed/trained for the specific classifier on the left.
Worst means the larger of FGSM-curr and AttNet errors for each $\eta$. 
%Note that FGSM fails to attack hardened networks (Adv FGSM40 and Minimax-Grad), 
%whereas AttNet can still attack them successfully.
}
\label{tbl:attnet_cifar10}
%\end{center}
\end{table}

\begin{table}[htb]
\renewcommand{\arraystretch}{0.9}
\small
\centering
%\begin{center}
\begin{tabular}{|c||cc|c|cc|c|}
\hline
{Defense\textbackslash Attack} & FGSM-curr & AttNet-curr & worst & FGSM-curr & AttNet-curr & worst\\
\hline
\if0
& \multicolumn{3}{|c|}{$\eta$=0.01} & \multicolumn{3}{|c|}{$\eta$=0.02} \\
\cline{2-7}
Minimax-AttNet& 0.583 & 0.213 & 0.583 & 0.613 & 0.221 & 0.613\\
Alt-AttNet    & 0.552 & 0.218 & 0.552 & 0.624 & 0.211 & 0.624\\
Minimax-Grad  & 0.342 & 0.212 & 0.342 & 0.390 & 0.242 & 0.390\\
\hline
& \multicolumn{3}{|c|}{$\eta$=0.03} & \multicolumn{3}{|c|}{$\eta$=0.04} \\
\cline{2-7}
Minimax-AttNet& 0.657 & 0.218 & 0.657 & 0.697 & 0.221 & 0.697\\
Alt-AttNet    & 0.680 & 0.221 & 0.680 & 0.702 & 0.231 & 0.702\\
Minimax-Grad  & 0.439 & 0.275 & 0.439 & 0.462 & 0.326 & 0.462\\
\hline
\fi
& \multicolumn{3}{|c|}{$\eta$=0.05} & \multicolumn{3}{|c|}{$\eta$=0.06} \\
\cline{2-7}
Minimax-AttNet & 0.731 & 0.239 & 0.731 & 0.733 & 0.248 & 0.733 \\
Alt-AttNet     & 0.721 & 0.238 & 0.721 & 0.743 & 0.255 & 0.743 \\
Minimax-Grad   & 0.481 & 0.343 & 0.481 & 0.453 & 0.484 & 0.484 \\
\hline
& \multicolumn{3}{|c|}{$\eta$=0.07} & \multicolumn{3}{|c|}{$\eta$=0.08} \\
\cline{2-7}
Minimax-AttNet & 0.762 & 0.256 & 0.762 & 0.775 & 0.266 & 0.775 \\
Alt-AttNet     & 0.743 & 0.257 & 0.732 & 0.771 & 0.258 & 0.771 \\
Minimax-Grad   & 0.487 & 0.529 & 0.529 & 0.462 & 0.607 & 0.607 \\
\hline
\end{tabular} 
\caption{Test error rates of Minimax-AttNet, Alt-AttNet, and Minimax-Grad
classifiers for CIFAR-10. 
Worst means the larger of FGSM-curr and AttNet errors for each row. 
%Minimax-AttNet is better than Alt-AttNet and Minimax-Grad against AttNet
%attacks, but worse than Minimax-Grad against the out-of-class attack (FGSM-curr).
}
\label{tbl:minimax_vs_others_cifar10}
%\end{center}
\end{table}

\if0
\begin{figure*}[thb]
\centering
%This is a placeholder.
\includegraphics[width=0.85\linewidth]{{figures/robustnet_catmouse_eta0.30_cifar10}.pdf}
\caption{Cat and mouse game of FGSM attacks and adversarial training for CIFAR-10.
The lower orange points are the error rates after adversarial training, 
and the upper green points are the error rates after FGSM attack. 
After 160 iterations ($\eta=0.3$), the error rate is still oscillating.
}
\label{fig:cat-and-mouse_cifar10}
\end{figure*}
\fi

\begin{figure*}[thb]
\centering
%This is a placeholder.
\includegraphics[width=1\linewidth]{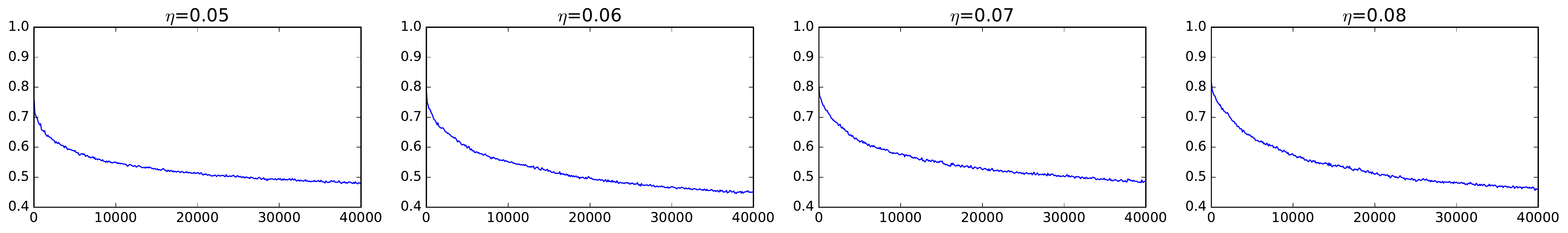}
\caption{Convergence of test error rates for Minimax-Grad with CIFAR-10.
}
\label{fig:normtrain_cifar10}
\end{figure*}

\begin{figure*}[thb]
\centering
%This is a placeholder.
\includegraphics[width=0.99\linewidth]{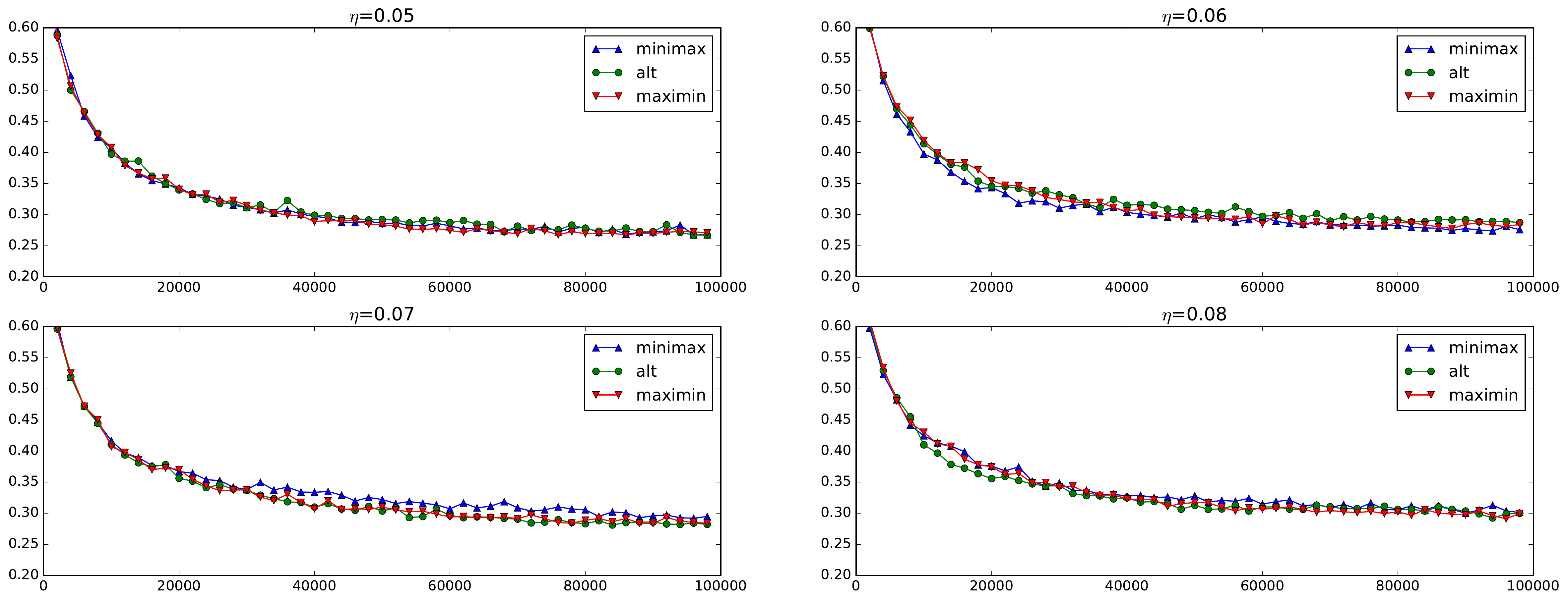}
\caption{Convergence of the test error rates for Minimax-AttNet (blue),
Alt-AttNet (green), and Maximin-AttNet (red) for CIFAR-10.
}
\label{fig:minimax_cifar10}
\end{figure*}

\end{document}